\documentclass[sigconf]{acmart}
\usepackage{times}
\usepackage{epsfig}
\usepackage{graphicx}
\usepackage{amsmath}
\usepackage{multirow}
\usepackage{booktabs}
\usepackage{subfigure}
\usepackage{threeparttable}
\renewcommand\footnotetextcopyrightpermission[1]{}
\usepackage{balance}
\AtBeginDocument{%
  }

\setcopyright{acmcopyright}
\copyrightyear{2023}
\acmYear{2023}
\acmDOI{XXXXXXX.XXXXXXX}

\acmConference[MM' 23]{Conference}{October 29, 2023}{Ottawa, Canada}
\acmPrice{15.00}
\acmISBN{978-1-4503-XXXX-X/18/06}

\acmSubmissionID{2243}

\hypersetup{
    colorlinks=true,
    linkcolor=red,
    urlcolor=red,
}

\begin{document}

\title{PVG: Progressive Vision Graph for Vision Recognition}


\author{Jiafu Wu}
\authornote{Both authors contributed equally to this research.}
\email{wujf21@m.fudan.edu.cn}
\orcid{0000-0002-1036-5076}
\affiliation{%
  \institution{School of Computer Science, Shanghai Key Laboratory of Data Science, Fudan University}
  \city{Shanghai}
  \country{China}
}

\author{Jian Li}
\authornotemark[1]
\email{swordli@tencent.com}
\orcid{0000-0002-0242-6481}
\affiliation{%
  \institution{Tencent Youtu Lab}
  \city{Shanghai}
  \country{China}
}

\author{Jiangning Zhang}
\author{Boshen Zhang}
\email{vtzhang@tencent.com}
\email{boshenzhang@tencent.com}
\orcid{0000-0001-8891-6766}
\orcid{0000-0001-9204-5676}
\affiliation{%
  \institution{Tencent Youtu Lab}
  \city{Shanghai}
  \country{China}
}

\author{Mingmin Chi}
\authornote{Corresponding authors.}
\email{mmchi@fudan.edu.cn}
\orcid{0000-0003-2650-4146}
\affiliation{%
  \institution{School of Computer Science, Shanghai Key Laboratory of Data Science, Fudan University}
  \institution{Zhongshan PoolNet Technology Ltd}
  \city{Shanghai}
  \country{China}
}

\author{Yabiao Wang}
\authornotemark[2]
\email{caseywang@tencent.com}
\orcid{0000-0002-6592-8411}
\affiliation{%
  \institution{Zhejiang University}
  \institution{Tencent Youtu Lab}
  \country{China}
}

\author{Chengjie Wang}
\email{jasoncjwang@tencent.com}
\orcid{0000-0003-4216-8090}
\affiliation{%
  \institution{Tencent Youtu Lab}
  \city{Shanghai}
  \country{China}
}

\renewcommand{\shortauthors}{Trovato and Tobin, et al.}

\newcommand{\wjf}[1]{\textcolor{red}{#1}}

\begin{abstract}
  \textit{Convolution-based} and \textit{Transformer-based} vision backbone networks process images into the grid or sequence structures, respectively, which are inflexible for capturing irregular objects. Though Vision GNN (ViG) adopts graph-level features for complex images, it has some issues, such as inaccurate neighbor node selection, expensive node information aggregation calculation, and over-smoothing in the deep layers. To address the above problems, we propose a \textbf{P}rogressive \textbf{V}ision \textbf{G}raph (PVG) architecture for vision recognition task. Compared with previous works, PVG contains three main components: \textit{1)} Progressively Separated Graph Construction (PSGC) to introduce second-order similarity by gradually increasing the channel of the global graph branch and decreasing the channel of local branch as the layer deepens; \textit{2)} Neighbor nodes information aggregation and update module by using Max pooling and mathematical Expectation (MaxE) to aggregate rich neighbor information; \textit{3)} Graph error Linear Unit (GraphLU) to enhance low-value information in a relaxed form to reduce the compression of image detail information for alleviating the over-smoothing. Extensive experiments on mainstream benchmarks demonstrate the superiority of PVG over state-of-the-art methods, \textit{e.g.}, our PVG-S obtains 83.0\% Top-1 accuracy on ImageNet-1K that surpasses GNN-based ViG-S by +0.9$\uparrow$ with the parameters reduced by 18.5\%, while the largest PVG-B obtains 84.2\% that has +0.5$\uparrow$ improvement than ViG-B. Furthermore, our PVG-S obtains +1.3$\uparrow$ box AP and +0.4$\uparrow$ mask AP gains than ViG-S on COCO dataset.\wjf{\url{https://github.com/wujiafu007/PVG/tree/main}}
\end{abstract}

\maketitle

\section{Introduction}
The Convolutional Neural Network (CNN)~\cite{lecun1998gradient} has been the dominant approach for general Computer Vision (CV) tasks~\cite{li2019dsfd, li2021asfd} due to its powerful inductive bias for capturing local information~\cite{simonyan2014very, krizhevsky2017imagenet, szegedy2015going, he2016deep}. However, the recent introduction of the transformer with attention mechanism from Natural Language Processing (NLP) into CV by ViT~\cite{dosovitskiy2020image} has shown that the vision transformer can achieve better performance than CNN with a large amount of pre-training. Subsequent ViT-based works~\cite{liu2021swin, wang2021pyramid, liu2022swin, dong2022cswin, li2022uniformer, zhang2022eatformer, zhang2021analogous, zhang2023rethinking} further demonstrate the powerful capabilities of the transformer in various downstream tasks, such as object detection~\cite{wu2023towards} and segmentation~\cite{Li2022SFNetFA, han2023referencetwice}. The studies have expedited the trend of unifying the Transformer architecture in the fields of NLP and CV.

\begin{figure}[]
 {\centering
 \includegraphics[width=0.48\textwidth]{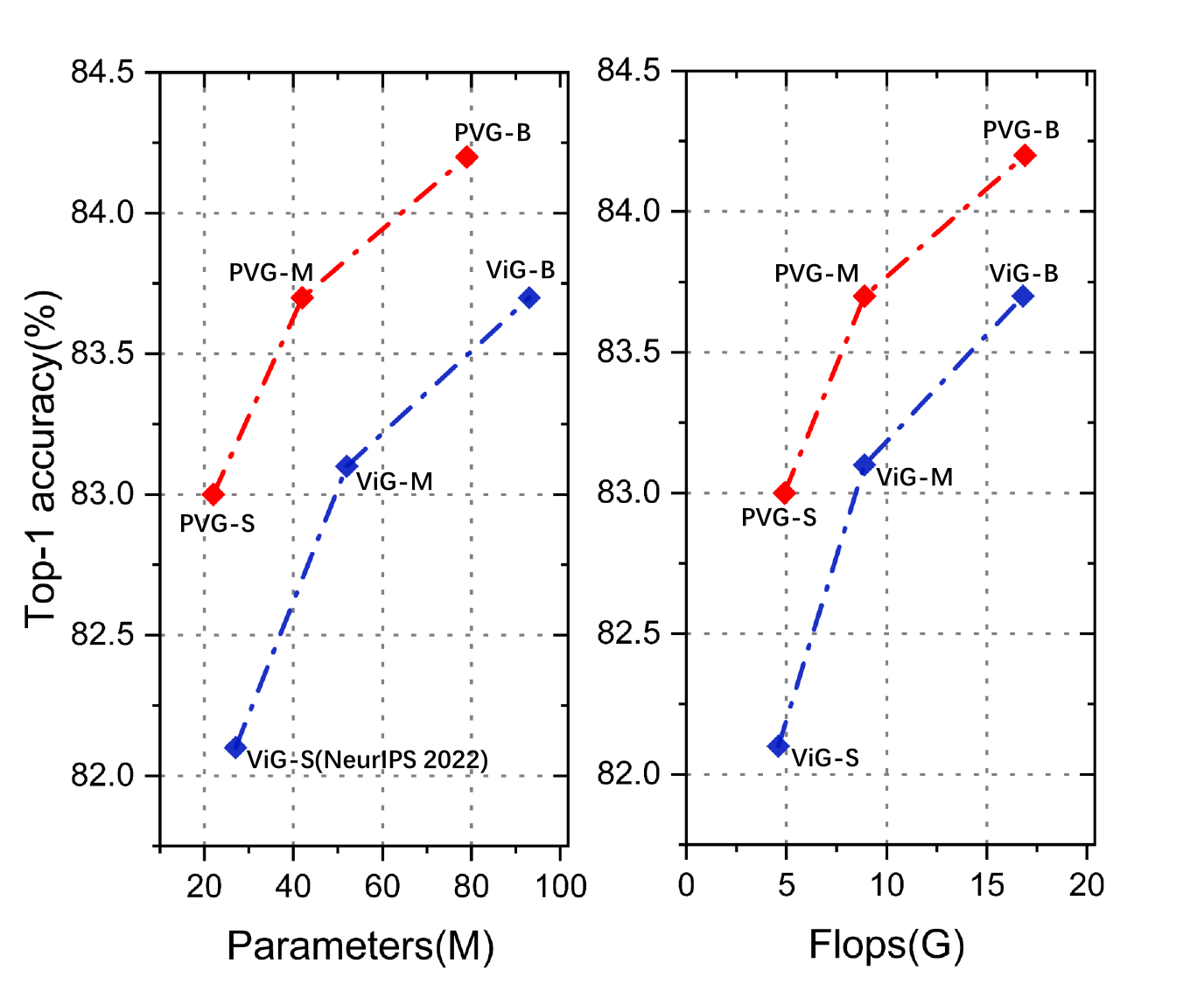}}
 \vspace{-2em}
	\caption{The comparison of Accuracy vs. Parameters and FLOPs between PVG and ViG. Our PVG achieves the best balance between accuracy and computation on the Imagenet.}
	\label{fig: the comparison of accuracy}
\end{figure}

However, the above CNNs and Transformers are designed to handle regular grid data (e.g., images) or sequential data (e.g., text) and are not flexible enough to capture non-Euclidean graph-structured data. To address this challenge, researchers have proposed a number of graph neural networks (GNN) technologies~\cite{bruna2013spectral, defferrard2016convolutional, levie2018cayleynets, niepert2016learning, velivckovic2017graph, monti2017geometric}. Since real-world objects often have irregular structures and various sizes, images can be viewed as a graph composition of parts. Some advanced graph-based approaches~\cite{chen2019multi, you2020cross, yang2020dpgn, ma2019gcan, roy2021curriculum} have been proposed for visual recognition. In addition, various GNN methods have been proposed for various downstream tasks, such as point cloud~\cite{li2019deepgcns}, object detection~\cite{zhao2021graphfpn}, and segmentation~\cite{wu2020bidirectional}. Moreover, some graph-based works~\cite{chen2019graphflow, chen2020iterative, chen2019reinforcement, wang2020amr} achieve competitive results in language processing tasks. Given these advancements, it is reasonable to ask whether GNN can serve as the backbone of computer vision and become a strong contender for the unified architecture of image, language, and graph-structured data in the future.

ViG~\cite{han2022vision} is the pioneering work in the field of computer vision that uses Vision Graph Neural Network (GNN) to process images. The model divides the input image into several patches and treats each patch as a node to enable information interaction. Despite significant progress, ViG still faces several challenges that remain unaddressed. ViG constructs graphs for images using only simple first-order similarity, which can result in inaccurate selection and connections among neighbor nodes. For graph information interaction, ViG applies two advanced Graph node information update functions in the GNN field: MR GraphConv~\cite{li2019deepgcns}, which has the lowest computational burden but poor performance, and EdgeConv~\cite{wang2019dynamic}, which performs well but has a high computational burden. Therefore, a node aggregation and an update function that can balance computation and performance well is needed. Additionally, ViG still experiences over-smoothing issues~\cite{li2019deepgcns} in deep layers. As the GNN goes deeper, the node representations become similar or even identical after certain layers, leading to a loss of information and reduced performance.

Towards the problems above, in this paper, we propose a new vision GNN architecture called Progressive Vision Graph (PVG) for general-purpose vision tasks. Fig.~\ref{fig:block transfor} illustrates our PVG structure design, which follows previous pyramid principles~\cite{wang2021pyramid} and is designed in a cascaded four-stage manner, with each stage composed of several graph blocks. Fig.~\ref{fig:block transfor} also highlights three key contributions of our PVG. Our first contribution is Progressively Separated Graph Construction (PSGC), which captures crucial second-order similarity information without increasing computation cost. PSGC splits the vision graph into three graphs constructed from different similarity measures in the channel dimension. In the shallow layer, we allocate a small part of the channel to the global graph. As the layer deepens, the channel ratio of the global graph gradually increases, fully absorbing the local information obtained from the local branch. This process calculates the second-order similarity of the nodes' features, which is equivalent to the similarity of their neighbors. PSGC is the first method to generalize the second-order similarity~\cite{wang2016structural}, an important concept in the GNN field, to vision tasks. The whole process is computationally efficient and does not introduce additional costs. We also propose a new node information aggregation and update method called MaxE, which is based on the distribution of neighbor representations. MaxE consists of three parts: node self-representation, the mathematical expectation of neighbor nodes, and the maximum Euclidean distance between the central node and its neighbor. Our experiments show that the proposed MaxE is effective and efficient, achieving the same performance as the powerful EdgeConv~\cite{wang2019dynamic} while reducing the number of parameters by 70\%. To address the over-smoothing problem, we propose GraphLU, a simple yet effective method that regenerates the node features to increase their diversity at deep layers with minimal computation and memory cost. GraphLU is a concise activation function that can be generalized to various computer vision backbone networks. Our proposed PVG achieves significant performance gains compared to the advanced graph-based backbone ViG~\cite{han2022vision}, while reducing the model parameters by around 20\%. Fig.~\ref{fig: the comparison of accuracy} illustrates the performance improvement of PVG over VIG. To summarise, the main contributions of this paper are as follows:

\begin{itemize}
 \item PSGC method introduces the important second-order similarity concept from the GNN field into the computer vision, providing more accurate neighbor node selection;
 \item A novel graph node aggregation and update mechanism MaxE to obtains state-of-the-art performance and low computation cost for vision tasks;
 \item An improved concise activation function GraphLU to solve the over-smoothing in the deep layers of Vision GNN without any addition computation cost;
 \item Our proposed PVG accelerates the generalization progress of Graph Neural Network (GNN) to Computer Vision (CV).
\end{itemize}

\section{Related Work}
\label{sec:typestyle}

\subsection{CNN and Transformer for Vision} 
Convolutional neural networks (CNNs) have been highly successful in modeling local dependencies, making them the leading approach in computer vision since the introduction of AlexNet~\cite{krizhevsky2017imagenet}. Over time, researchers have developed increasingly effective CNN architectures, including VGG~\cite{simonyan2014very}, GoogleNet~\cite{szegedy2015going}, ResNet~\cite{he2016deep}, MobileNet~\cite{howard2017mobilenets}, and others, which have further propelled the field forward. 

\begin{figure*}[]
 {\centering
 \includegraphics[width=0.95\textwidth]{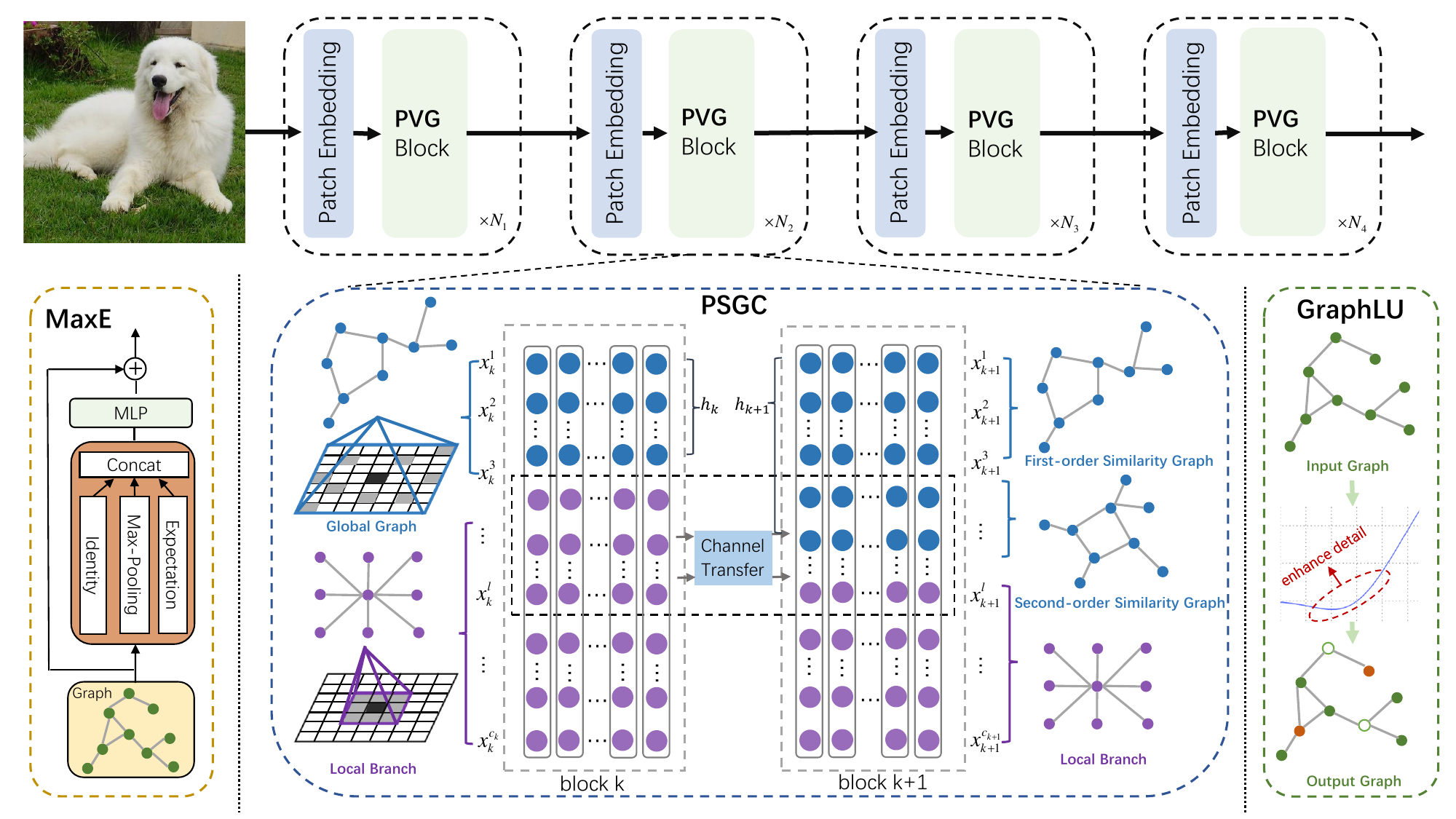}}
	\caption{Our PVG architecture is designed in a cascaded four-stage manner, with each stage adopting the Progressively Separated Graph Construction (PSGC) to introduce second-order similarity, which transfers channels from local to global graphs between adjacent blocks. After graph construction, PVG uses our proposed MaxE in each block for information aggregation and update. Additionally, PVG utilizes a concise activation function GraphLU to enhance detail for alleviating the over-smoothing problem.}
	\label{fig:block transfor}
\end{figure*}

However, the introduction of ViT~\cite{dosovitskiy2020image} has challenged the dominance of CNNs in computer vision. Subsequent works, such as Deit~\cite{touvron2021training}, PVT~\cite{wang2021pyramid}, T2T~\cite{yuan2021tokens} and Swin~\cite{liu2021swin}, have achieved impressive performance in various downstream vision tasks. Additionally, some research efforts have proposed novel hybrid structures, such as CvT~\cite{wu2021cvt}, Coatnet~\cite{dai2021coatnet}, Container~\cite{gao2021container}, and ViTAE~\cite{xu2021vitae}, which leverage the strengths of both CNN and Transformer models. These efforts elevate the field of computer vision to a higher level. Recently, some works make attempts to strike back at Transformer architecture in computer vision by designing advanced pure CNN architectures, such as ConvNeXt~\cite{liu2022convnet} and RepLKNet~\cite{ding2022scaling}, which show competitive results across various vision tasks and are faster than Transformer models in inference.

\subsection{Graph Neural Network}
There are two main branches of graph neural network research for non-European graph data: frequency domain graph convolution~\cite{bruna2013spectral, henaff2015deep, defferrard2016convolutional, kipf2016semi, levie2018cayleynets, li2018adaptive} and spatial domain convolution~\cite{atwood2016diffusion, niepert2016learning, hamilton2017inductive, velivckovic2017graph, monti2017geometric, huang2018adaptive}. Representative works that improve GNN performance by enabling convolution operations on non-Euclidean datasets, including ChebNet~\cite{defferrard2016convolutional}, GCNs\cite{kipf2016semi}, GraphSAGE~\cite{hamilton2017inductive} and GAT~\cite{velivckovic2017graph}. Furtherly, DeepGCNs~\cite{li2019deepgcns} introduces concepts from CNNs such as residual structure and dilated convolution, into GNNs, realizing a 56-layer graph convolution network. Subsequent works including Mixhop~\cite{abu2019mixhop}, GDC~\cite{gasteiger2019diffusion}, and S2GC~\cite{zhu2021simple}, have further extended the depth and generalization capabilities of GNNs for graph data such as social network and biochemical graphs.

GNNs have been extensively applied in computer vision tasks, such as few-shot learning and zero-shot learning~\cite{chen2019multi, you2020cross, yang2020dpgn, kampffmeyer2019rethinking, liu2021isometric}. They have also facilitated advancements in domain adaptation and generalization~\cite{ma2019gcan, roy2021curriculum, luo2020progressive, chen2022compound}. In addition, some works have also utilized GNNs to enhance the features of the CNN backbone in object detection tasks~\cite{zhao2021graphfpn, li2022sigma, zhang2020dynamic}, perform graph reasoning on class pixels~\cite{hu2020class}, and design bidirectional graph reasoning networks for panoramic segmentation~\cite{wu2020bidirectional}. Few works have directly utilized GNNs as the backbone network for computer vision. ViG~\cite{han2022vision} is the only such work, which uses patches from the input image as nodes and dynamically selects nodes with the top first-order similarity as neighbors. Despite these contributions, ViG still faces several challenges that remain
unaddressed, such as inaccurate neighbor node selection, expensive
node information aggregation and update calculation, and over-smoothing in the
deep layers. 

\subsection{Over-smoothing Problem of GNN}
The over-smoothing~\cite{li2018deeper} problem in GNNs is typically described as node features becoming excessively similar or even identical after a certain layer in the model, leading to a decline in model performance. Some works~\cite{li2019deepgcns, xu2018representation} propose improvements on GNN structures to alleviate the over-smoothing issue. The essence of structural methods is to increase the nonlinear connections in the model to enrich the node features, but it may increase the computational cost. In addition, some works propose a variety of regularization methods~\cite{rong2019dropedge, zhao2019pairnorm} to address the over-smoothing problem. Furthermore, there are various other methods proposed in the literature, including MixHop~\cite{abu2019mixhop}, which aims to increase the receptive field of GNNs, and GDC~\cite{gasteiger2019diffusion}, which leverages graph diffusion for regularization. While these methods are effective for non-Euclidean graph data, they have poor generalization performance, and directly applying them to visual tasks often leads to performance degradation. In addition to adopting an FFN module like VIG for node feature transformation and encouraging node diversity, Our PVG also utilizes a concise activation function GraphLU to alleviate the over-smoothing.

\section{Method}
\label{sec:exp}

\subsection{Recap the concept of Vision Graph}
We introduce the abstract concept of the Vision Graph, which refers to a general structure for representing visual information. Specifically, the input data is initially embedded into a sequence of nodes:
\begin{align}
    H = NodeEmbedding\left( X \right).
    \label{equ:equ1}
\end{align}
The sequence of nodes ${H^{\left( {N \times C} \right)}}$ is subsequently inputted into a sequence of graph blocks:
\begin{align}
    H' = H + f\left( {g\left( {Norm\left( H \right)} \right)} \right)
    \label{equ:equ2}
\end{align}
\begin{align}
    H'' = H' + FFN\left( {Norm\left( {H'} \right)} \right),
    \label{equ:equ3}
\end{align}
where $g$ is a graph structure learning function that constructs images into graphs; $f$ is the node information aggregation and update function after constructing the graphs; FFN is the feed-forward network.

\subsection{First-order and Second-order Similarity}
\begin{theorem}
\textbf{First-order Similarity.} The first-order similarity between nodes "i" and "j" can be expressed as follows:
\begin{align}
    S_{ij}^1 = l\left( {{v_i},{v_j}} \right)
    \label{equ:equ4}
\end{align}
where l represents a distance metric such as Euclidean distance, cosine distance, or dot product distance.
\label{thm-1}
\end{theorem}

\begin{theorem}
\textbf{Second-order Similarity.} The second-order similarity between nodes "i" and "j" in a graph can be expressed as follows:
\begin{align}
    S_{ij}^2 = l\left( {\varphi (N({v_i})),\varphi (N({v_j}))} \right)
    \label{equ:equ5}
\end{align}
where $N({v_i})$ represents the local neighbors of node i, and $\varphi$ is the information aggregation function such as summation, averaging, or other suitable methods.
\label{thm-1}
\end{theorem}

\subsection{Progressively Separated Graph Construction}
Compared to fully-connected transformers, accurately describing the similarity between nodes is crucial when constructing a graph, as the connections in the graph are sparse. Missing essential connections or connecting noisy edges can significantly harm the model's performance. A straightforward approach is to calculate the first-order similarity between nodes and subsequently remove unimportant connection edges through a specific sparsification method.
\begin{align}
    {{\vec A}_{i,:}} = topk\left( {{{S_{i,:}}} } \right)
    \label{equ:equ7}
\end{align}

\begin{figure}[]
 {\centering
 \includegraphics[width=0.49\textwidth]{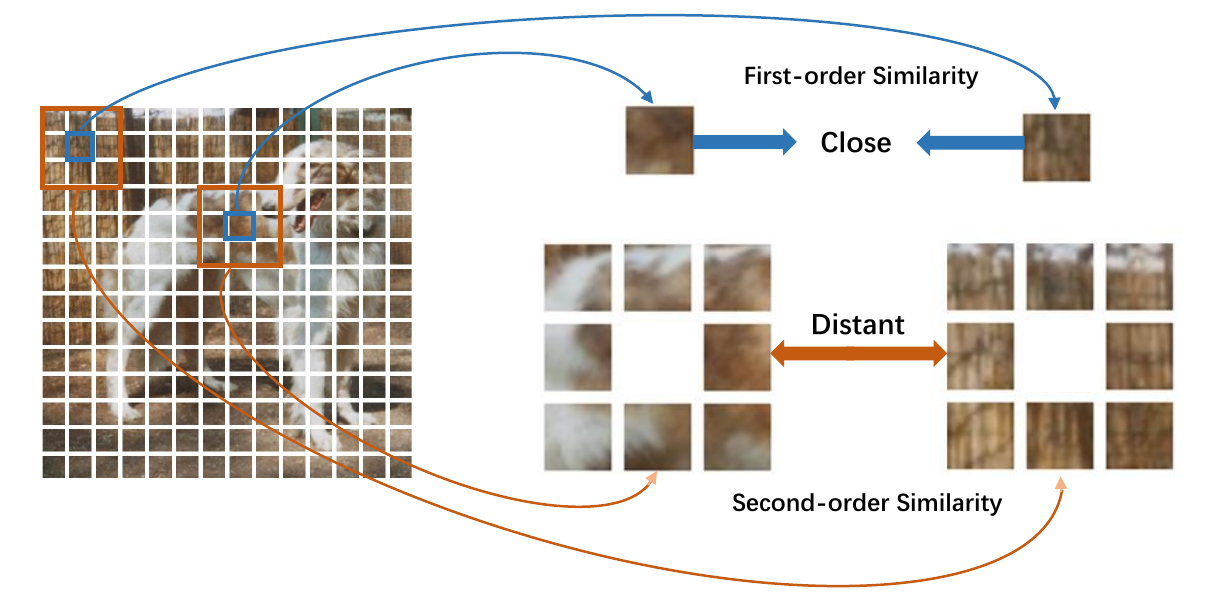}}
	\caption{Illustration of First-order similarity and Second-order similarity.}
	\label{fig2}
\end{figure}

However, the first-order similarity measure may not be sufficient to accurately capture complex inter-node relationships. As illustrated in Figure~\ref{fig2}, the first-order similarity metric exhibits a bias towards low-level features such as color and texture, leading the model to select noisy neighbor nodes. In contrast, the second-order similarity is more semantic, as it describes the feature similarity of the central node's neighbors in the Euclidean space. Second-order similarity can effectively compensate for the limitations of first-order similarity by rewarding connections between nodes that are similar to both themselves and their neighbors while penalizing the noisy nodes that are similar to themselves but dissimilar to their surroundings. This results in the model selecting more precise and semantically relevant node connections under sparse, limited connectivity.

To introduce the second-order similarity without additional computational cost, we propose a Progressively Separated Graph Construction called PSGC. As shown in Figure~\ref{fig:block transfor}, the PSGC consists of two branches. The local branch has more channels to capture the contextual information of each node in the shallow layer. The other branch constructs the graph globally by computing the first-order similarity between nodes. Mathematically, we denote the input sequence of nodes as $x = \left( {{x_1}, \ldots ,{x_n}} \right)$, and the output of the graph nodes $\tilde x = \left( {{{\tilde x}_1}, \ldots ,{{\tilde x}_n}} \right)$, where ${x_i},{{\tilde x}_i} \in {R^c}$. The computation of our PVG block could be formulated as:
\begin{align}
    {{\tilde x}_i} = Concat\left[ {f\left( {{S_{ij}},x_{i}^{{h_k}},x_{j}^{{h_k}}} \right),\mathop \sum \limits_{j = 1}^n {\alpha _{ij}}x_{j}^{C - {h_k}}}\right]
    \label{equ:equ9}
\end{align}
where $f$ is the nodes' information aggregation and update function of the global graph. In addition, we set a distance threshold to the local branch, specifically set it to 0 if the Chebyshev distance of node $i$ and $j$ is greater than a threshold $r$($r$ defaults to 3). 

As the model goes deeper, the proportion of local branch channels reduces, and the local branch gradually transfers a portion of its channels to the global graph. These channels have sufficiently absorbed local information in the shallow layers. Therefore, when the global graph computes the first-order similarity between nodes on these channels, it is equivalent to computing the second-order similarity between the nodes, which can be expressed as follows:
\begin{align}
    S_{ij}^2 = \mathop \sum \limits_l^{{h_{k + 1}} - {h_k}} \left( {\mathop \sum \limits_t^{{N_i}} {\alpha _{it}}x_t^l \cdot \mathop \sum \limits_t^{{N_j}} {\alpha _{jt}}x_t^l} \right)
    \label{equ:equ8}
\end{align}
where $h_k$ denotes the channels in the kth layer of the global graph and $x_t^l$ denotes the neighbor of node i or node j in the local branch.

PVG ultimately manifests as a trident structure, consisting of a local branch, a first-order similarity graph, and a second-order similarity graph. These three branches represent three distinct inductive biases in visual tasks: the local branch represents local dependency, the first-order similarity graph represents global dependency, and the second-order similarity graph represents a mixed dependency between local and global. The channel allocation ratio of the three branches can be controlled by adjusting the number of channels allocated to the progressive separation operation. However, we posit that all three structures at different levels are necessary, and the absence of any one of them would lead to a decrease in model capacity.

\subsection{MaxE on node representation learning}
After constructing the graph, it is necessary to design a nodes information update mechanism. Similar to CNNs, GNNs' update process of node information at the \textit{l}-th layer can be abstractly described as follows:
\begin{align}
    x_i^{l + 1} = \phi \left( {x_i^l,\rho \left( {\{ x_j^l|x_j^l \in {\rm N}\left( {x_i^l} \right)\} ,x_i^l,{W_1}} \right),{W_2}} \right)
    \label{equ:equ10}
\end{align}
where ${{\rm N}\left( {x_i^l} \right)}$ represents the set of neighbors of node \textit{i} at the \textit{l}-th layer. $\rho $ is a node feature aggregation function and $\phi $ is a node feature update function. ${{W_1}}$ and ${{W_2}}$ are learnable matrices. 

The difficulty of the node aggregation and update function $\phi $ lies in how to sample limited nodes from a given set of neighboring nodes to sufficiently represent the information of the entire set, as full sampling is expensive. We propose a concise yet comprehensive sampling strategy for the $\phi $, which is formulated as follows:
\begin{align}
    x_i^{l + 1} = W \cdot Concat\left[ {x_i^l,\max \left( {x_j^l - x_i^l} \right),mean(x_j^l)} \right]
    \label{equ:equ11}
\end{align}
It consists of three parts: self identity map, the maximum pooling of the difference between self and neighbor nodes, and the mathematical expectation of neighbor nodes. In other words, for a given set of neighboring nodes, the neighborhood information could be fully expressed by only sampling two points: the mean point and the point with the maximum difference from the central node. 

To better understand the working mechanism of Formula~\ref{equ:equ11},  we need to have a detailed discussion on the key term $\rho $.
\begin{align}
    \rho \left(  \cdot  \right) = \max \left( {x_j^l - x_i^l|x_j^l \in {\rm N}\left( {x_i^l} \right)} \right)
    \label{equ:equ12}
\end{align}
Another common form of $\rho $ in the field of GNN is as follows:
\begin{align}
    \rho \left(  \cdot  \right) = \frac{1}{m} {\mathop \sum \limits_{j = 1}^m \left( {x_j^l - x_i^l|x_j^l \in {\rm N}\left( {x_i^l} \right)} \right)} 
    \label{equ:equ13}
\end{align}
Setting ${z_1} = {x_j^l} - {x_i^l}$, then we define the following:
\begin{align}
    z_1^\prime  = \max \left( {{z_1}} \right), {{\bar z}_1} = mean\left( {{z_1}} \right)
    \label{equ:equ14}
\end{align}
\begin{align}
    z_1^{\prime \prime } = \mathop {\arg \max }\limits_{{z_{1j}}} \left( {z_1^\prime  - {z_{1j}}} \right)
    \label{equ:equ15}
\end{align}
Then Formula~\ref{equ:equ12} can be expressed as:
\begin{align}
    \rho \left(  \cdot  \right) & = \max \left( {{z_1}} \right) \\
    & = {{\bar z}_1} + \left( {z_1^\prime  - {{\bar z}_1}} \right)  \\
    & = {{\bar z}_1} + \left( {z_1^{\prime \prime } - {{\bar z}_1}} \right) + \left( {z_1^\prime  - z_1^{\prime \prime }} \right)  \\
    & = {{\bar z}_1} + \left( {z_1^{\prime \prime } - {{\bar z}_1}} \right) + \max \left( {z_1^\prime  - {z_{1j}}} \right)
    \label{equ:equ17}
\end{align}

$\rho \left(  \cdot  \right)$ is divided into three parts: the first item is ${{\bar z}_1}$, which implies that Formula~\ref{equ:equ13} is a subset of Formula~\ref{equ:equ12}; the second part is $\left( {z_1^{\prime \prime } - {{\bar z}_1}} \right)$, which tends to be small so we call it the remainder; the third part is $\max \left( {z_1^\prime  - {z_{1j}}} \right)$, which we call the maximum bound within the class. 
\begin{figure}[htb]
 {\centering
 \includegraphics[width=0.45\textwidth]{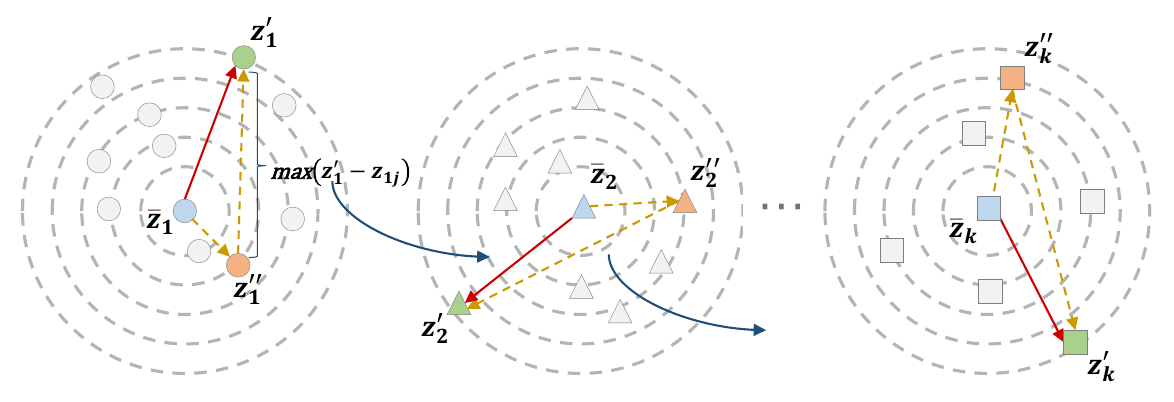}}
	\caption{Illustration of neighbor sampling in MaxE.}
	\label{fig:sample}
\end{figure}
We provide an illustration of the above ideas, as shown in Figure~\ref{fig:sample}. In particular, if we let ${z_2} = z_1^\prime - {z_{1j}} $ for the third item above, there will be the following iterative form:
\begin{align}
    \rho '\left(  \cdot  \right) & = \max \left( z_1^\prime - {{z_{1j}} } \right) \\
    & = {{\bar z}_2} + \left( {z_2^{\prime \prime } - {{\bar z}_2}} \right) + \max \left( {z_2^\prime  - {z_{2j}}} \right)
    \label{equ:equ18}
\end{align}
If k times are iterated, there is the following k-order feature aggregation function $\rho$:
\begin{align}
    {\rho ^{k - 1}}\left(  \cdot  \right) = {{\bar z}_k} + \left( {z_k^{\prime \prime } - {{\bar z}_k}} \right) + \max \left( z_k^\prime- {{z_{kj}} } \right)
    \label{equ:equ19}
\end{align}
The final form of aggregate function $\rho$ can be expressed in the following abstract form:
\begin{align}
    \rho \left(  \cdot  \right) = [{{\bar z}_1} +  \cdots  + {{\bar z}_k}] + \tilde C + {\rho ^k}\left(  \cdot  \right)
    \label{equ:equ19}
\end{align}
where ${z_1} = {x_j^l} - {x_{\scriptstyle i \hfill \atop 
  \scriptstyle  \hfill}^l }$, ..., ${z_2} = z_1^\prime - {z_{1j}} $; ${\tilde C}$ is the accumulated remainder; ${\rho ^k}\left(  \cdot  \right)$ is the k-order feature aggregation function. 

Formula~\ref{equ:equ19} is a form similar to the Taylor expansion, indicating that max pooling sampling iteratively retains the information of neighboring nodes at each order in an iterative manner. However, it still leaves out the zero-order neighbor information ${z_0} = {x_j^l}-0$. For the completeness of information at each order, we add $mean(z_0)$ in the final Formula~\ref{equ:equ11}. We refer to the final form as MaxE, it is a sufficiently rich sampling scheme for neighbor information. Adequate comparative experiments show that MaxE outperforms existing state-of-the-art graph representation learning methods for visual tasks.

\subsection{GraphLU towards over-smoothing}
In ViG~\cite{han2022vision}, over-smoothing phenomenon still appears in the deep layers. One intuitive solution is to add a non-linear transformation layer to enrich the node representation. However, this operation would increase computational overhead and model parameters. Therefore, we aimed to alleviate the over-smoothing problem with almost no increase in computation, and designed a new general activation function called Graph Linear Units (GraphLU).

\begin{figure}[htb]
 {\centering
 \includegraphics[width=0.49\textwidth]{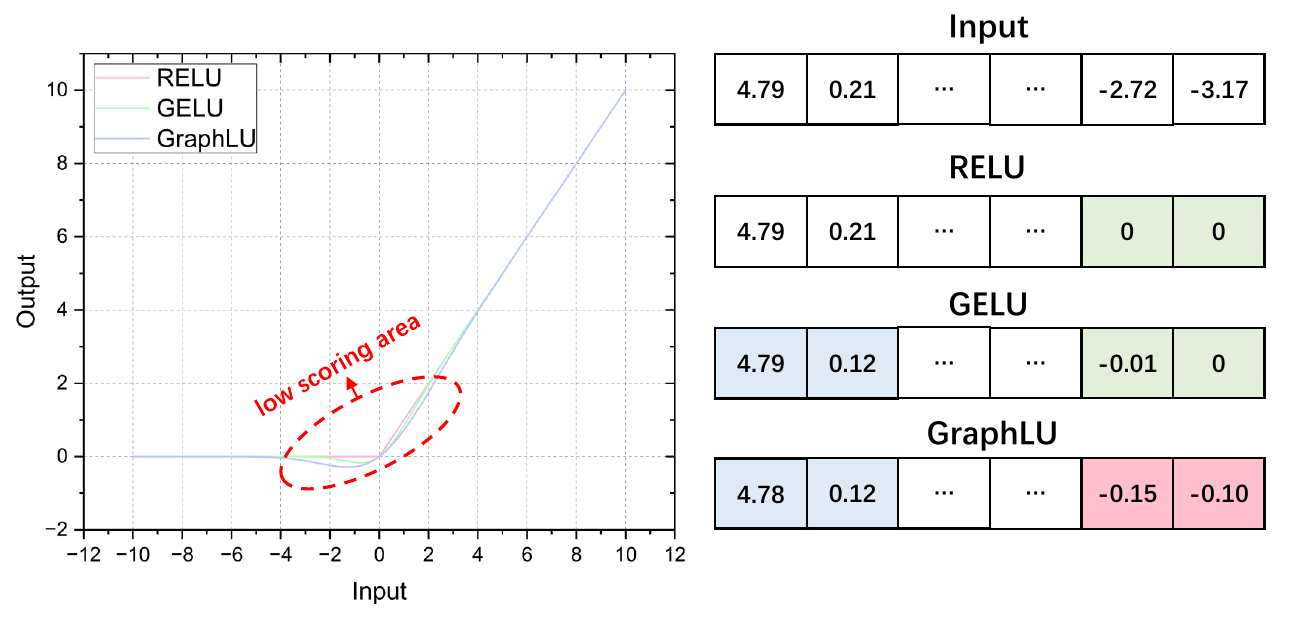}}
	\caption{Illustration of GraphLU ($\varepsilon = 1$) for alleviating the over-smoothing problem of graph network.}
	\label{fig:GraphLU}
\end{figure}

Firstly we review Relu~\cite{hahnloser2000digital}, from which we will derive the GraphLU. Relu is a zero-or-identity mapping, it performs an identity output for the high-value part of the input x, and zeros the low-value part:
\begin{align}
    \operatorname{Re} LU\left( x \right) = \max \left( {0,x} \right)
    \label{equ:equ3}
\end{align}

Our main idea is that the ReLU activation function discards some "unimportant" parts of node representations, which results in a reduction of the node representation space and causes all outputs to converge to the same representation at deep layers. We think that these discarded "unimportant" low-value information still contribute to the richness of the final node representations, as they often contain detailed information. Retaining some of these low-scoring information can significantly enrich the node representation. GELU~\cite{hendrycks2016bridging} is another activation function that retains some low-scoring information and has shown SOTA performance in Transformer models. We further discuss the potential of this relaxation operation by introducing a random variable $X_{ij}$.
\begin{align}
    {X_{ij}} \sim  N\left( {0,{{\left( {1 + \varepsilon } \right)}^2}} \right),{X_{ij}} \in {R^{m \times n}}
    \label{equ:equ3}
\end{align}
where m is the batch size , n is the number of nodes and $\varepsilon $ a learnable parameter. We hope that ${X_{ij}}$ with large values have a greater probability of being retained, but at the same time retain some ${X_{ij}}$ with small values hovering around 0. Assuming the input is ${x_{ij}}$, we can achieve this by defining the following probability:
\begin{align}
    \phi \left( {{x_{ij}}} \right) = p\left( {{X_{ij}}  \le {x_{ij}}} \right)
    \label{equ:equ3}
\end{align}

That means, for an input ${x_{ij}}$, we describe the probability that the random variable ${X_{ij}}$ has a value less than its value, which can be found by integrating:
\begin{align}
    \phi \left( {{x_{ij}}} \right) = \mathop \smallint \nolimits_{ - \infty }^{{x_{ij}}} \frac{1}{{\sqrt {2\pi } \left( {1 + \varepsilon } \right)}}{{\text{e}}^{ - \frac{{{t^2}}}{{2{{\left( {1 + \varepsilon } \right)}^2}}}}}{\text{d}}t
    \label{equ:equ3}
\end{align}
An approximate expression is the Gauss Error function (erf), which is well supported in various deep learning frameworks, and the mathematical description is as follows:
\begin{align}
    erf\left( \xi  \right) = \frac{2}{{\sqrt \pi  }}\mathop \smallint \nolimits_0^\xi  {{\text{e}}^{ - {t^2}}}{\text{dt}}
    \label{equ:equ3}
\end{align}
Let $t = \frac{z}{{\sqrt 2 (1 + \varepsilon )}}$,then:
\begin{align}
     erf\left( \xi  \right) = \frac{2}{{\sqrt {2\pi } \left( {1 + \varepsilon } \right)}}[\mathop \smallint \nolimits_{ - \infty }^{\sqrt 2 (1 + \varepsilon )\xi } {{\text{e}}^{ - \frac{{{z^2}}}{{2{{\left( {1 + \varepsilon } \right)}^2}}}}}{\text{dz - }}\phi \left( 0 \right)]
    \label{equ:equ3}
\end{align}
\begin{align}
     2\phi \left( {\sqrt 2 (1 + \varepsilon )\xi } \right) = erf\left( \xi  \right) + 2\phi \left( 0 \right)
    \label{equ:equ3}
\end{align}
where $\phi \left( 0 \right) = \frac{1}{2}$, let $x = \sqrt 2 (1 + \varepsilon )\xi $, we can get the final form of $\phi (x)$:
\begin{align}
     \phi (x) = \frac{1}{2}erf\left[ {\frac{x}{{\sqrt 2 (1 + \varepsilon )}} + 1} \right]
    \label{equ:equ3}
\end{align}
Multiply the input x by the probability $\phi (x)$ to get the final form of GraphLU:
\begin{align}
     GraphLU(x) = 0.5x \cdot erf\left[ {\frac{x}{{\sqrt 2 (1 + \varepsilon )}} + 1} \right]
    \label{equ:equ3}
\end{align}

The above derivation follows the principle of GELU, but it reveals a key insight: by just adding a small relaxation perturbation $\varepsilon$ to the variance of the random variable $X_{ij}$, we can elegantly amplify detail information and enrich node features without introducing additional computational cost, thus alleviating over-smoothing issues.

\section{Experiments}
To demonstrate the effectiveness of PVG as a computer vision backbone, we conduct classification experiments on ImageNet1K~\cite{deng2009imagenet}, object detection experiments on COCO~\cite{lin2014microsoft}. In addition, we conduct comprehensive ablation experiments on the three main components in PVG to demonstrate that each part brings tangible improvements to the model. Furthermore, we provide some visualization experimental results to support our point of view.

\subsection{Evaluation On IMAGENET-1K}
For the fairness of the experiment,the input image size is 224x224,and we follow the training strategy in DeiT~\cite{touvron2021training}, using the same data augmentations and regularization methods. Specifically, we use AdamW optimizer, and init learning rate $1 \times {10^{ - 3}}$ with via cosine decaying. Furthermore, in order to support the generalization of downstream tasks and the stability of training, we use LePE~\cite{dong2022cswin} as the position embedding and use LayerScale~\cite{zhu2021gradinit} in the last two layers.

We compare our PVG with those representative vision networks in Table~\ref{tab:all_results}, our PVG series can  be comparable to the Transformer variants, and outperform the state-of-the-art CNN networks. In particular, compared with ViG~\cite{han2022vision}, which is also based on GNN, PVG has a significant lead in Top1-accuracy.
\begin{table}[htb]
\caption{Comparison of different models on ImageNet-1K.}
\begin{threeparttable}
\resizebox{0.48\textwidth}{6cm}{
\begin{tabular}{l|c|ccc|c}
\toprule
Model  & Mixing Type  & Resolution & \#Param.            & FLOPs      & Top-1    \\
       \midrule
ResNet-50~\cite{wightman2021resnet}   & Conv & 224$\times$224 & 26M  & 4.1G  & 79.8    \\
ConvNeXt-T~\cite{liu2022convnet}   & Conv & 224$\times$224 & 29M  & 4.5G  & 82.1    \\
PVT-S~\cite{wang2021pyramid}   & Attn & 224$\times$224 & 25M  & 3.8G  & 79.8    \\
T2T-ViT-14~\cite{yuan2021tokens}   & Attn & 224$\times$224 & 22M  & 4.8G  & 81.5   \\
Swin-T~\cite{liu2021swin}   & Attn & 224$\times$224 & 29M  & 4.5G  & 81.3   \\
ViL-S~\cite{li2021localvit}   & Attn & 224$\times$224 & 25M  & 4.9G  & 81.8    \\
Focal-T~\cite{yang2021focal}   & Attn & 224$\times$224 & 29M  & 4.9G  & 82.2    \\
CrossFormer-S~\cite{wang2023crossformer++}   & Attn & 224$\times$224 & 31M  & 4.9G  & 82.5    \\
RegionViT-S~\cite{chen2021regionvit}   & Attn & 224$\times$224 & 31M  & 5.3G  & 82.6    \\
ViG-S~\cite{han2022vision} & Graph & 224$\times$224 & 27M & 4.6G  & 82.1    \\
\midrule
PVG-S(ours) & Graph & 224$\times$224 & 22M & 5.0G  & \textbf{83.0}    \\
\midrule
ResNet-101~\cite{wightman2021resnet}   & Conv & 224$\times$224 & 45M  & 7.9G  & 81.3    \\
ConvNeXt-S~\cite{liu2022convnet}   & Conv & 224$\times$224 & 50M  & 8.7G  & 83.1    \\
PVT-M~\cite{wang2021pyramid}   & Attn & 224$\times$224 & 44M  & 6.7G  & 81.2   \\
T2T-ViT-19~\cite{yuan2021tokens}   & Attn & 224$\times$224 & 39M  & 8.5G  & 81.9  \\
Swin-S~\cite{liu2021swin}   & Attn & 224$\times$224 & 50M  & 8.7G  & 83.0     \\
Focal-S~\cite{yang2021focal}   & Attn & 224$\times$224 & 51M  & 9.1G  & 83.5     \\
CrossFormer-B~\cite{wang2023crossformer++}   & Attn & 224$\times$224 & 52M  & 9.2G  & 83.4    \\
RegionViT-M~\cite{chen2021regionvit}   & Attn & 224$\times$224 & 41M  & 7.4G  & 83.1    \\
ViG-M~\cite{han2022vision} & Graph & 224$\times$224 & 52M & 8.9G  & 83.1     \\
\midrule
PVG-M(ours) & Graph & 224$\times$224 & 42M & 8.9G  & \textbf{83.7}     \\
\midrule
ResNet-152~\cite{wightman2021resnet}   & Conv & 224$\times$224 & 60M  & 11.5G  & 81.8    \\
ConvNeXt-B~\cite{liu2022convnet}   & Conv & 224$\times$224 & 89M  & 15.4G  & 83.8    \\
PVT-L~\cite{wang2021pyramid}   & Attn & 224$\times$224 & 61M  & 9.8G  & 81.7    \\
T2T-ViT-24~\cite{yuan2021tokens}   & Attn & 224$\times$224 & 64M  & 13.8G  & 82.3     \\
Swin-B~\cite{liu2021swin}   & Attn & 224$\times$224 & 88M  & 15.4G  & 83.5      \\
ViL-B~\cite{li2021localvit}   & Attn & 224$\times$224 & 56M  & 13.4G  & 83.2    \\
Focal-B~\cite{yang2021focal}   & Attn & 224$\times$224 & 90M  & 16.0G  & 83.8       \\
CrossFormer-L~\cite{wang2023crossformer++}   & Attn & 224$\times$224 & 92M  & 16.6G  & 84.0   \\
RegionViT-B~\cite{chen2021regionvit}   & Attn & 224$\times$224 & 73M  & 13.0G  & 83.2    \\
ViG-B~\cite{han2022vision} & Graph & 224$\times$224 & 93M & 16.8G  & 83.7      \\
\midrule
PVG-B(ours) & Graph & 224$\times$224 & 79M & 16.9G  &  \textbf{84.2}     \\
\bottomrule
\end{tabular}
} 
\end{threeparttable}
\label{tab:all_results}
\end{table}

\begin{figure*}[htb]
\begin{minipage}[]{0.25\textwidth}
    \centering
    \centerline{\includegraphics[width=3.8cm]{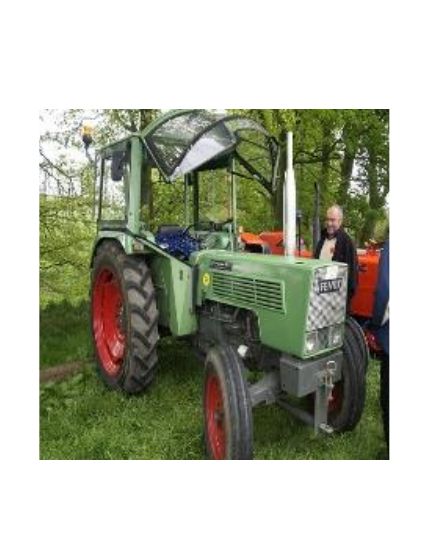}}
    \centerline{(a) Input image}
\end{minipage}
\begin{minipage}[]{0.355\textwidth}
    \centering
    \centerline{\includegraphics[width=5.46cm]{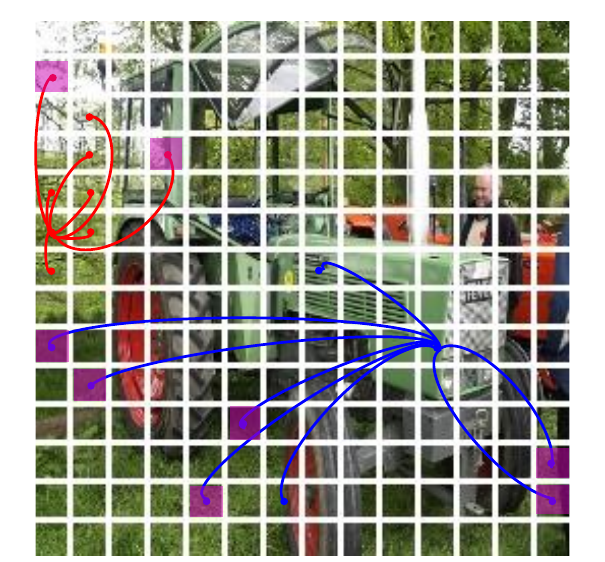}}
    \centerline{(b) ViG}
\end{minipage}
\begin{minipage}[]{0.355\textwidth}
    \centering
    \centerline{\includegraphics[width=5.65cm]{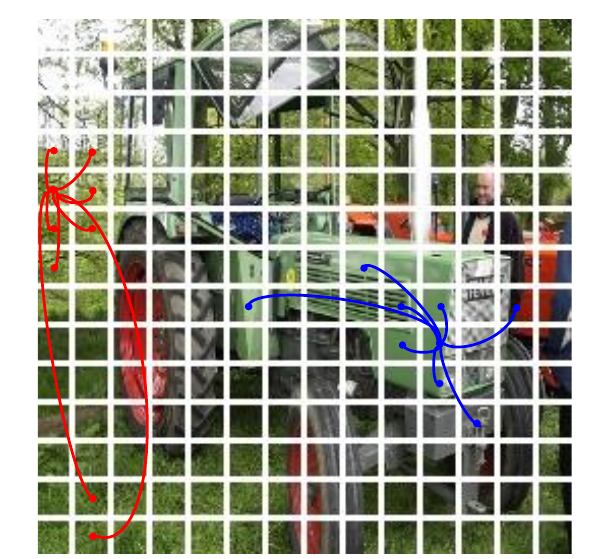}}
    \centerline{(c) PVG(ours)}
\end{minipage}
\caption{Due to its monotonous first-order similarity measure, ViG finds some unreasonable neighbor nodes(the part marked in purple). Our PVG employs a hybrid similarity measure, which finds neighbors that are more accurate and focused.}
\label{fig:visual}
\end{figure*}

To better understand how our PVG model works, we visualize the constructed graph structure in PVG-S and compare it with ViG-S. In Figure~\ref{fig:visual}, we show a graph sample in the 12th blocks. Two center nodes are visualized as drawing all the edges will be messy. We can observe that ViG, lacking second-order similarity information, erroneously connects the grass node and the truck node (highlighted in purple) due to their similar green color. However, since our PVG employs a more complex hybrid similarity measure, it does not solely rely on low-level features such as color or texture to select neighboring nodes. Consequently, the neighbors of center nodes in PVG are more semantic and belong to the same category.

\subsection{Evaluation On COCO}
We conduct experiments on the COCO detection dataset to evaluate the generalization of PVG. We pretrain the backbones on the ImageNet-1K dataset and follow the finetuning used in Swin Transformer~\cite{liu2021swin} on the COCO dataset. The models are trained in the commonly-used “1x” schedule with 1280×800 input size. 

\begin{table}[htb]
\caption{Object detection and instance segmentation results.}
\centering
\resizebox{0.48\textwidth}{1.6cm}
{
\begin{tabular}{c|cc|ccc|ccc}
\toprule

  & \multicolumn{8}{|c}{MASK R-CNN 1$\times$ Schedule}  \\
\cline{2-9}
    & \#Param(M)  & FLOPs(G)  & $A{P^b}$   & $AP_{50}^b$      & $AP_{75}^b$    & $A{P^m}$   & $AP_{50}^m$      & $AP_{75}^m$\\
       \midrule
ResNet-50~\cite{wightman2021resnet}   & 44.2 & 260.1 & 38.0  & 58.6  & 41.4   & 34.4  & 55.1  & 36.7 \\
PVT-S~\cite{wang2021pyramid} & 44.1 & \textbf{245.1} & 40.4  & 62.9  & 43.8   & 37.8  & 60.1  & 40.3 \\
Swin-T~\cite{liu2021swin}   & 47.8 & 264.0 & 42.2  & 64.6  & 46.2   & 39.1  & 61.6  & 42.2 \\
CycleMLP-B2~\cite{chen2021cyclemlp}   & 46.5 & 249.5 & 42.1  & 64.0  & 45.7   & 38.9  & 61.2  & 41.8 \\
ViG-S~\cite{han2022vision}   & 45.8 & 258.8 & 42.6  & 65.2  & 46.0   & 39.4  & 62.4  & 41.6 \\
\midrule
PVG-S(ours) &\textbf{40.9}  &267.9  &\textbf{43.9}   &\textbf{66.3}   &\textbf{48.0}    &\textbf{39.8}   &\textbf{62.8}   &\textbf{42.4}   \\
\bottomrule
\end{tabular}
}
\label{tab:coco_results}
\end{table}

\begin{figure*}[htb]
\begin{minipage}[]{1\textwidth}
    \centering
    \centerline{\includegraphics[width=15.8cm]{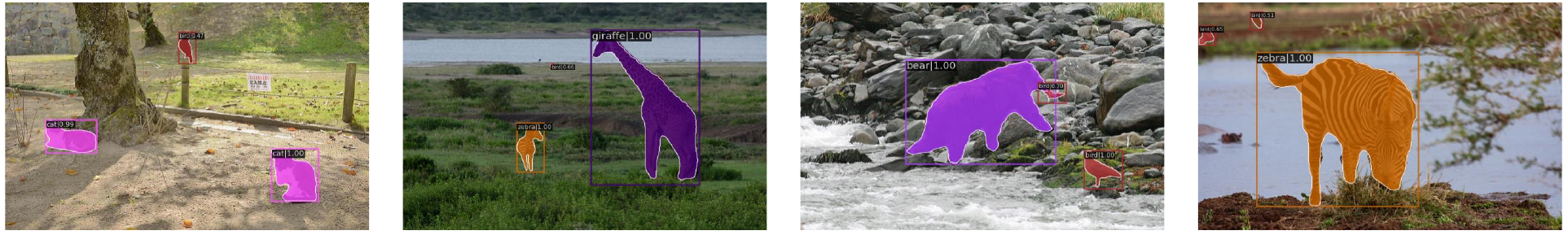}}
    \centerline{(a) Sparse objects}
\end{minipage}
\begin{minipage}[]{1\textwidth}
    \centering
    \centerline{\includegraphics[width=15.8cm]{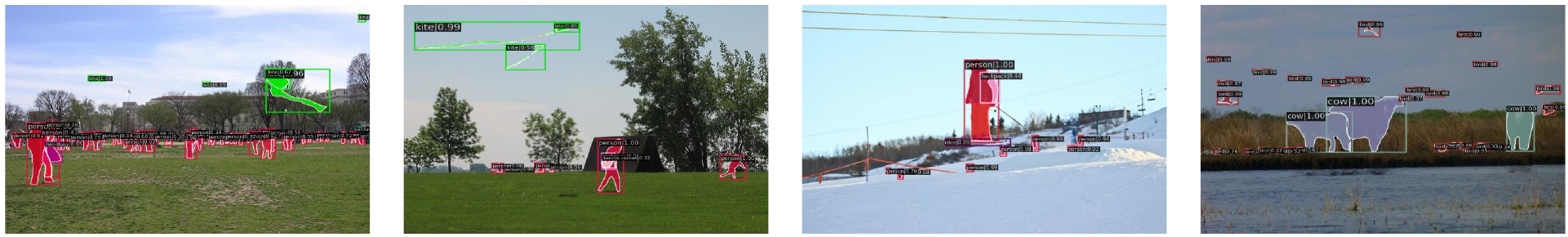}}
    \centerline{(b) Dense objects}
\end{minipage}
\begin{minipage}[]{1\textwidth}
    \centering
    \centerline{\includegraphics[width=15.8cm]{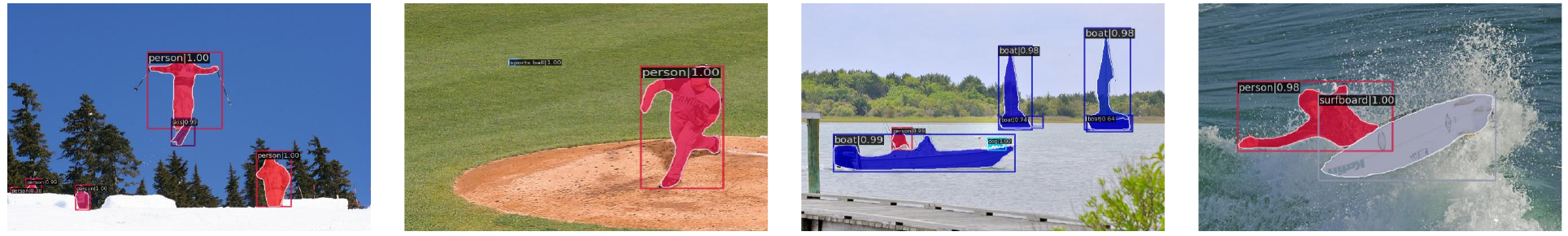}}
    \centerline{(c) Moving objects}
\end{minipage}
\caption{Qualitative examples for object detection and instance segmentation on PVG-S+Mask R-CNN. Our PVG achieves good detection
and instance segmentation performance on sparse objects, dense
objects, and moving objects.}
\label{fig:detection}
\end{figure*}

Table~\ref{tab:coco_results} compares our PVG with state-of-the-art GNN-based and Transformer-based architecture. The experimental results demonstrate that our PVG achieves further improvements in performance in object detection and can compete with other mainstream architectures. Additionally, Figure~\ref{fig:detection} visualizes some results of object detection and instance segmentation and achieves good detection and instance segmentation performance on sparse objects, dense objects, and moving objects.

\subsection{Ablation Study}
In order to thoroughly examine the effectiveness of different PVG methods, we meticulously design a series of ablation experiments under a completely fair setting. Each experiment strictly adheres to the same architecture and hyperparameters, with only one component altered per ablation experiment. The comprehensive results of these ablation experiments are presented in Table~\ref{tab:ablation2}. The findings clearly demonstrate the substantial impact of introducing second-order similarity information in the graph structure, as it significantly improves the overall performance. Additionally, our proposed novel graph node aggregation and update function, MaxE, as well as the newly designed activation function aimed at addressing over-smoothing, both contribute to further enhancing the model's performance without introducing any additional computational cost.
\begin{table}[htb]
\caption{Ablation experiments of main methods in PVG.}
\begin{threeparttable}
\centering
\resizebox{0.477\textwidth}{1.45cm}{
\begin{tabular}{c|c|c|c}
\toprule
Architecture design   & Params(M)  & FLOPs(G)    & Top1(\%)    \\
\midrule
Baseline (ViG-S layout)    & 27 & 4.6 & 81.9\tnote{*}    \\
+PSGC for second-order similarity  & 22 & 4.9 & 82.5(\textcolor{blue}{+0.6})  \\
MR GraphConv$ \to $MaxE  & 22 & 4.9 & 82.8(\textcolor{blue}{+0.3}) \\
GELU$ \to $GraphLU  & 22 & 4.9 & 83.0(\textcolor{blue}{+0.2})  \\
\bottomrule
\end{tabular}
}
\begin{tablenotes}    
        \footnotesize               
        \item[*] denotes the reproduced result in our experimental settings.
\end{tablenotes}
\end{threeparttable}
\label{tab:ablation2}
\end{table}

In order to demonstrate the superiority of our designed nodes' information update function MaxE, Table 4 conducts comparative experiments under the same settings. Compared with the two advanced methods in vision GNN, our MaxE has a best balance between accuracy and parameters, which achieve the accuracy improvements from 84.3 of MR GraphConv to 85.0. The benchmark unit of the parameters in the table is the parameters in GIN. (Due to limited computing resources, the data is a randomly selected subset of Imagenet, with a training set of 60,000 images and 150 categories.)

\begin{table}[htb]
\centering
\caption{Comparison of MaxE with other aggregate functions.}
\resizebox{0.475\textwidth}{0.85cm}{
\begin{tabular}{cccccc}
\toprule
       & GIN   & MR GraphConv & EdgeConv    & GraphSAGE      & MaxE(ours)           \\
       \hline
Top1-acc    & 80.55    & 84.3 & 85.0    & 83.76  & \textbf{85.0}     \\
\midrule
Params   & 1    & 2 & 10    & 11   & \textbf{3}     \\
\bottomrule
\end{tabular}
}
\label{Aggregate function comparison}
\end{table}


\subsection{Over-smoothing analysis}
\begin{figure}[]
 {\centering
 \includegraphics[width=0.49\textwidth]{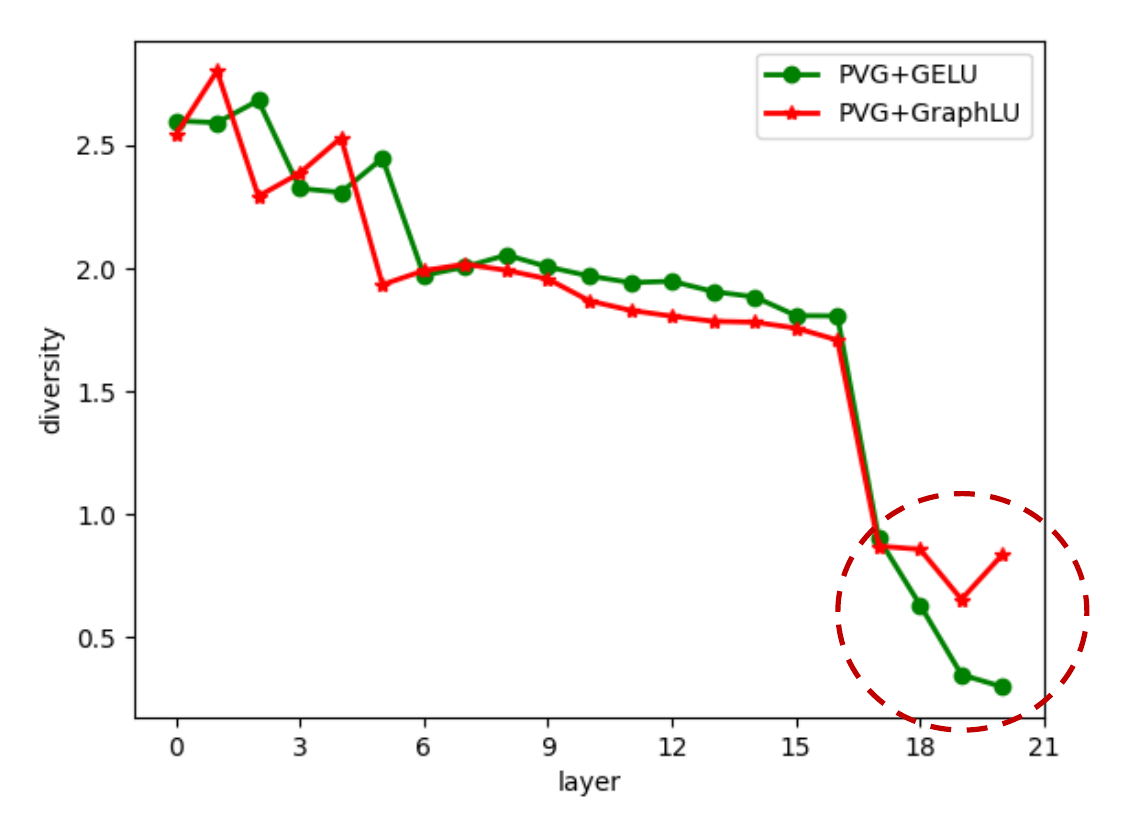}}
	\caption{GraphLU alleviates the over-smoothing phenomenon in the deep layer without increasing the computational cost.}
	\label{fig:oversmooth}
\end{figure}
We use Imagenet-1K to train two small PVG models with 21 layers under the exact same setting. The only difference between them is that one uses GELU as the activation function, and the other uses our GraphLU. As shown in Figure~\ref{fig:oversmooth}, the node features distinctiveness of PVG using GELU drops sharply from layer 17 to layer 21 until the last layer is almost 0, where $diversity = \frac{1}{n} \sum \limits_{i = 1}^n \left \Vert {x_i -  \frac{1}{n} \sum \limits_{i = 1}^n x_i  }  \right \Vert_2$. However, the over-smoothing phenomenon of PVG with GraphLU has been alleviated. The node features distinctiveness only decreases once, and stay at an acceptable level in the last few layers instead of dropping to 0. Notably, GraphLU achieves a certain level of alleviation of the over-smoothing issue, albeit not as pronounced as in some other methods. Importantly, this improvement is attained with minimal computational overhead. Furthermore, it outperforms any previous endeavors dedicated to tackling the over-smoothing problem, exhibiting notably higher training efficiency.

\section{Conclusion}
\label{sec:prior}
In this work, we explore graph-based backbone network design for vision and introduce a novel structure named Progressive Vision Graph (PVG). PVG constructs two separate graphs for global and local modeling and a fast graph node aggregation and update mechanism MaxE for neighbor nodes. In addition, we propose a concise activation function to effectively solve the over-smoothing phenomenon in previous work, which allows the vision GNN network to be stacked to 20 layers or even deeper to deal with various complex vision tasks. Extensive experiments on image recognition and object detection demonstrate the superiority of the proposed PVG architecture compared with previous state-of-the-art works. In summary, our proposed PVG accelerates the generalization process of graph neural networks towards computer vision tasks. This advancement positions GNN as a compelling contender for the unified architecture of image, language, and graph-structured data in the future.

\section*{ACKNOWLEDGEMENT}
\label{ACKNOW}
This work was supported in part by Natural Science Foundation of China under contract 62171139, and in part by Zhongshan science and technology development project under contract 2020AG016.



\bibliographystyle{ACM-Reference-Format}
\balance
\bibliography{sample-base}






\end{document}